\documentclass[10pt, a4paper]{article}

\usepackage[
]{lrec-coling2024} 
\usepackage[ruled,vlined]{algorithm2e} 
\usepackage{amsmath}
\usepackage{float}
\usepackage{graphicx}
\usepackage[]{todonotes}
\usepackage{balance}
\usepackage[hang, flushmargin]{footmisc} 

\title{Digital Guardians:\\Can GPT-4, Perspective API, and Moderation API reliably detect hate speech in reader comments of German online newspapers?
\thanks{\bf Disclaimer:
This research aims to combat hate speech and, therefore, contains examples of hate speech or offensive language, for analysis and educational purposes.}}

\name{Manuel Weber, Moritz Huber, Maximilian Auch,\\ {\bf \large Alexander Döschl*}, {\bf \large  Max-Emanuel Keller}, {\bf \large  Peter Mandl}} 

\address{\\HM Hochschule München University of Applied Sciences \\
         Lothstraße 64, 80335 München, Germany \\
         *Corresponding author: alexander.doeschl@hm.edu\\}

\abstract{
In recent years, toxic content and hate speech have become widespread phenomena on the internet. Moderators of online newspapers and forums are now required, partly due to legal regulations, to carefully review and, if necessary, delete reader comments. This is a labor-intensive process. Some providers of large language models already offer solutions for automated hate speech detection or the identification of toxic content. These include GPT-4o from OpenAI, Jigsaw’s (Google) Perspective API, and OpenAI’s Moderation API. Based on the selected German test dataset HOCON34k, which was specifically created for developing tools to detect hate speech in reader comments of online newspapers, these solutions are compared with each other and against the HOCON34k baseline. The test dataset contains 1,592 annotated text samples. For GPT-4o, three different promptings are used, employing a Zero-Shot, One-Shot, and Few-Shot approach. The results of the experiments demonstrate that GPT-4o outperforms both the Perspective API and the Moderation API, and exceeds the HOCON34k baseline by approximately 5 percentage points, as measured by a combined metric of MCC and F2-score.
 \\ \newline }

\begin{document}

\maketitleabstract

\section{Introduction}
\label{sec:intro}

Toxic content and hate speech are increasingly spreading across social media and other online platforms, including forums and comment sections \citep{udanor_combating_2019}. The digital space, which, according to \citet{jaki_hate_2023}, was originally characterized by the advantages of unfiltered communication, has since experienced a negative shift. The dissemination of harmful content online poses a significant challenge to both society and democracy. In response, legislators have implemented measures to counteract the spread of hate speech. As of February 2024, under the Digital Services Act \citep{european_parliament_verordnung_2022}, not only large online companies but also smaller online platforms operating within the EU are required to take effective action against illegal content, including hate speech. This calls for solutions that can assist in detecting and removing harmful content without infringing on freedom of expression.

Automated hate speech detection, powered by artificial intelligence, is one approach to addressing this issue. While most of the research on hate speech detection is concentrated on English, some studies are beginning to address other languages.
Pretrained Large Language Models (LLMs) like GPT-4 and available moderation APIs are potential tools for this task.

This study compares the effectiveness of OpenAI's GPT-4 \citep{openai_gpt-4_2023}, Google's Perspective API \citep{perspective_perspective_2024}, and OpenAI's Moderation API \citep{openai_openai_2024} in detecting hate speech in German online newspapers. The publicly accessible HOCON34k dataset \citep{keller_hocon34k_2024}, containing annotated reader comments from German online newspapers, serves as the basis for evaluation. Each solution uses a pretrained model, with no fine-tuning or additional training applied for this study. The datasets and models underlying these solutions, except for the baseline model, are not publicly disclosed. However, the companies usually provide paid API services for public use.

Various definitions of hate speech can be found in the literature (see, for example, \citet{eric_europaische_2016} and \citet{meta_hassrede_2024}). For this study, we use the comprehensive hate speech guidelines defined by \citet{keller_hocon34k_2024}, derived from the behavior codes of online newspapers. This definition includes the following types of content: racist and xenophobic content, sexist and homophobic content such as misogyny or misandry, hostility towards LGBTQ+ individuals, religious hatred, antisemitism, other forms of hate against humanity, unconstitutional or extremist slurs, vulgar, obscene, or offensive language, insults, threats, and harassment.   

\section{Related Work}
\label{sec:relwork}
Hate speech detection using machine learning methods has been extensively studied in numerous research projects in recent years. Initially, traditional machine learning methods dominated; however, with the introduction of the transformer architecture~\citep{vaswani_attention_2017}, deep learning techniques have taken the forefront. Transformer-based, pretrained language models, especially Large Language Models (LLMs), have become increasingly important, leading to notable progress in hate speech detection for English texts, as well as other languages \citep{istaiteh_racist_2020,alkomah_literature_2022,jahan_systematic_2023,rawat_hate_2024}.

For instance, \citet{chiu_detecting_2021} utilized GPT-3 for hate speech detection via Zero-, One-, and Few-Shot learning. Their findings indicate that Few-Shot learning improved performance by approximately 25\,\% compared to the Zero- and One-Shot approaches \citep{chiu_detecting_2021}. Similarly, \citet{guo_investigation_2024} used GPT-3.5 Turbo with different prompts in a Few-Shot learning context. An F1-score of 0.82 was reported for hate speech recognition in English texts. In comparison, the F1-score for Chinese texts was only 0.55. \citet{li_hot_2023} compared ChatGPT with MTurker annotations and report that ChatGPT achieves an accuracy of around 0.8 for malicious texts. \citet{matter_close_2024} evaluated the performance of GPT in recognizing violent speech on the platform \mbox{\emph{incels.is}}, using GPT to augment text examples. GPT-4 was found to outperform GPT-3.5 in all metrics, with a weighted F1-score of 0.88 and a macro F1-score of 0.78 reported as the best values.
\citet{pan_comparing_2024} conducted a comparison between fine-tuning pretrained BERT-based models and several Large Language Models (LLMs), including Mistral-7B-Instruct, Zephyr-7b-beta, and Tulu-2, for detecting sexist and misogynistic hate speech, as well as hate speech against migrants, using two datasets in English language. The LLMs were prompted using Zero- and Few-Shot learning strategies. The best result was achieved with 5-Shot learning and Zephyr, which yielded a macro F1-score of 0.7094 in detecting sexist speech. In comparison, the fine-tuned DeBERTa model achieved a macro F1-score of 0.8681. \citet{glasebach_gmhp7k_2024} accomplished a macro-average F1-Score up to 0.79 for hate speech and 0.75 for misogynistic hatespeech using a fine-tuned GBERT-base model~\citep{chan_germans_2020}.

Google Jigsaw developed the publicly available Perspective API \citep{google_jigsaw_perspective_2024}, which utilizes a proprietary pretrained model designed to detect harmful content, including sexist or racist speech, and threats. OpenAI also provides a currently free API, the Moderation API, for detecting harmful content across eleven categories (e.g., sexist text, harassment, hate, threats, etc.) \citep{openai_openai_2024}. The studies by \citet{markov_holistic_2022} evaluate OpenAI's Moderation API and Google's Perspective API with different datasets and report better performance of the Moderation API for all datasets except Google's Jigsaw dataset. For example, AUPRC values exceeding 0.9 were achieved using the Stormfront dataset, while the Perspective API yielded an AUPRC value above 0.87.
Additionally, experiments using GPT-2 showed better results than with the Perspective API. \citet{hosseini_deceiving_2017} identified limitations in the Perspective API when handling modified passages, while \citet{nogara_toxic_2023} reported that the Perspective API performed better for German content compared to other tools.
Previous studies have often not utilized the latest versions of GPT models or APIs and have primarily focused on hate speech detection in English. Our experiments aim to compare GPT-4o with Google Jigsaw's and OpenAI's APIs, based on their versions as of June 2024. The hate speech detection will be tested using a recently released German-language dataset specifically designed for detecting hate speech in online newspapers (HOCON34k) \citep{keller_hocon34k_2024}. Metrics such as the F2-score, Matthews correlation coefficient (MCC), and a combined metric of both will be used for comparison. A fine-tuned model based on Google BERT  \citep{devlin_bert_2019} developed by \citet{keller_hocon34k_2024} for hate speech detection in the HOCON34k dataset will serve as the baseline.    

\section{Methodology}
\label{sec:methodology}
\subsection{Objective and Experimental Setup}
The objective of our experiments was to compare the performance of GPT-4o, Perspective API, and Moderation API against the HOCON34k baseline for hate speech detection in a binary classification task:

\begin{itemize} 
\item For GPT-4o, prompts were used for Zero-Shot, One-Shot, and Few-Shot learning. The output was mapped to a binary classification ("Hate Speech" or "No Hate Speech" / "Yes" or "No"). 
\item For the Moderation API, the overall score of a query was used to derive the classification (True = "Hate Speech", False = "No Hate Speech"). 
\item For the Perspective API, the output value representing the probability of hate speech was mapped to a binary classification based on three different threshold values. 
\item The HOCON34k baseline utilizes a pre-trained BERT model~\citep{chan_germans_2020} fine-tuned with the HOCON34k dataset. The HOCON34k classifier outputs a probability score for hate speech, which is evaluated using an optimized threshold. 
\end{itemize}

In total, seven individual tests were conducted: three using GPT-4o, with varying learning approaches (abbreviated in Alg.~\ref{alg:experiment} as GPT4-Z/O/F), three using the Perspective API with varying threshold values $\tau \in \{0.38, 0.5, 0.8\}$ (denoted as PAPI-038/05/08 in Alg.~\ref{alg:experiment}), and one test using the Moderation API (MAPI). The results of the HOCON34k baseline were taken from \citet{keller_hocon34k_2024} for comparison.

The evaluation of detection performance was based on a uniform test dataset designed for binary classification with the labels \emph{Hate Speech} or \emph{No Hate Speech}. The test dataset was a selected portion of the HOCON34k dataset \citep{keller_hocon34k_2024}, annotated by twelve individuals, including professional moderators from online newspapers. The HOCON34k test dataset contains real-world examples from reader comments on various online newspaper platforms. For the GPT-4o tests, the hate speech definition from HOCON34k was inputted as a prompt. The Perspective API and Moderation API used their own definitions of hate speech.

\begin{algorithm}[ht]
\caption{Detection Experiment}
 \textbf{begin}\\
    \hspace{0.5em} MAX\_REPEATS := 3\\
    \hspace{0.5em} SAMPLES := 1592\\
    \hspace{0.5em} init api[] with \{GPT4-Z, GPT4-O, GPT4-F,\\
    \hspace{3em} PAPI-038, PAPI-05, PAPI-08, MAPI\}\\
    \hspace{0.5em} init result[api, MAX\_REPEATS, SAMPLES]\\
    \hspace{0.5em} init metrics[api, MAX\_REPEATS]\\
    \hspace{0.5em} init avgMetr[api]\\
    \hspace{0.5em}~\textbf{for}~{a in api}~\textbf{do}\\
        \hspace{2em} \textbf{for}~{i from 1 to MAX\_REPEATS}~\textbf{do}\\
            \hspace{3em} \textbf{for}~{t from 1 to SAMPLES}~\textbf{do}\\
                \hspace{4em} response = a.request(t.text)\\
                \hspace{4em} result(a, i, t) = computeForecast(\\\hspace{6em}
                threshold, response, t.label)\\
            \hspace{3.5em}\textbf{end for}\\
            \hspace{3.2em} compute metrics[a, i] $\leftarrow$ \mbox{result[a, i, *]}\\
        \hspace{2.5em}\textbf{end for}\\
        \hspace{2.2em} compute avgMetr[a] = avg(metrics[a, i])\\
    \hspace{1em}\textbf{end for}\\
    \hspace{0.7em} print avgMetr\\
\textbf{end}
\label{alg:experiment}
\end{algorithm}

For the comparison of the approaches, each individual test applied the 1,592 data samples to make predictions. All tests were repeated three times. Confusion matrices were generated, and the metrics Recall, Precision, F1-score, F2-score, MCC, and the Champion-Challenger score \(S\) \citep{keller_hocon34k_2024} were computed. An average value for each metric was calculated across the three runs. The three runs were also used to verify the consistency of the outputs. A Python script was developed to automate the API requests for all runs. Algorithm~\ref{alg:experiment} outlines its functionality in pseudocode. The result calculation was performed three times for each considered API. The results for each request were mapped to the confusion matrix using the \texttt{computeForecast} function.
Finally, the evaluation metrics are computed for each run and API, followed by calculating the average across the three runs.
The following subsections introduce the test dataset~(\ref{sec:dataset}), evaluation metrics~(\ref{sec:metrics}), and give further details on the experiments~(\ref{sec:GPT}-\ref{sec:PersAPI}).

\subsection{Dataset HOCON34k}
\label{sec:dataset}

The basis of our experiments is the HOCON34k dataset (Hatespeech in Online Comments from German Newspapers, comprising approximately 34,000 comments) from \citet{keller_hocon34k_2024}. The dataset originates from the comment sections of various German newspapers, including the \textit{TZ} and the \textit{Frankfurter Rundschau}. The texts were labeled as \emph{Hate Speech} or \emph{No Hate Speech} by professional moderators and other experts, including researchers and IT specialists. In total, 29 individuals participated in the annotation process. The complete dataset contains 34,223 text examples, of which 28,992 were labeled as not containing hate speech (84.7\,\%) and 5,231 as hate speech (15.3\,\%).

A subset of 15,248 texts, including 12,275 training samples, 2,492 validation samples, and 1,592 test samples, was selected for our comparison purposes. This subset, annotated by 12 experts under the supervision of a lead moderator, exhibited high quality with an inter-rater agreement of \( \kappa = 0.6078 \) (Fleiss’ Kappa), indicating substantial agreement.

For our experiments, only the test dataset is required, consisting of 329 (20.67\,\%) hate speech texts and 1,263 (79.33\,\%) non-hate speech texts. These 1,592 texts, representing around 10\,\% of the total dataset, were used for testing. In addition to the text and binary annotation, the dataset includes information such as post ID, annotator ID, whether the context was sufficient for annotation, and the source file name. However, for our experiments, only texts and labels were used while context information was not considered. Table 1 provides a few example entries from the original dataset.

\begin{table*}[!ht]
\begin{center}
\begin{tabular}{|l|p{3.5cm}|p{8cm}|}
    \hline
    \textbf{Post Id} & \textbf{Label} & \textbf{Text (Original and \textit{Translated})} \\
    \hline
    4235999044 & {Hate Speech \newline (enough context)}& Muslimischer Terrorismus. Wia immer hoid. \newline \textit{Translation: Muslim terrorism. As always.} \\
    \hline
    3248014695 & {Not Hate Speech \newline (enough context)}& Hkh vfkmfjhi \newline \textit{Translation: Hkh vfkmfjhi} \\
    \hline
    3811968397 & {Hate Speech \newline (not enough context)}& Wir werden nur verar... \newline \textit{Translation: We're just being p... }\\
    \hline
    4964167063 & {Not Hate Speech \newline (not enough context)}& Nein Es sind doch nicht alle betrunken Obwohl ? \newline \textit{Translation: No, they're not all drunk Although ?} \\
    \hline
\end{tabular}
\caption{Example texts from the HOCON34k dataset \citep{keller_hocon34k_2024}}
\end{center}
\end{table*}

The texts are primarily in German. In addition, some test data include foreign languages or internet links. Individual texts in the dataset contain spelling and grammar errors, colloquial and dialectal expressions, or simply unintelligible character strings. Symbols and self-censorship also appear in some of the comments.

\subsection{Evaluation Metrics}
\label{sec:metrics}

Accuracy, Precision, Recall, and F1-score are calculated according to~\citet{powers_evaluation_2020} based on the confusion matrix using the four values: True Positive (TP), True Negative (TN), False Positive (FP), and False Negative (FN).
The moderators involved in annotating the HOCON34k dataset primarily expect hate speech detection to minimize undetected hate speech, meaning FN should be as low as possible. A higher rate of false positives (FP) is tolerable in this context, as all user-generated comments undergo a manual review process. For this reason, recall is more important than precision in our use case. Consequently, in addition to the F1-score, the F2-score is particularly relevant, as it gives twice as much weight to recall compared to precision, as shown in equation~\ref{eq:F2}.

\begin{equation}
\label{eq:F2}
F_2 = \frac{(1+2)^2 \cdot TP}{(1+2)^2 \cdot TP + FP + 2^2 \cdot FN}
\end{equation}

However, an overly one-sided emphasis on recall should be avoided, which is why the Matthews Correlation Coefficient (MCC) is also considered. The MCC is defined in equation~\ref{eq:MCC} using $P$ and $N$ as the total number of positives~($P$) and negatives~($N$). The F2-score and a normalized MCC (eq.~\ref{eq:M_norm}) are combined into a single metric, referred to in \citet{keller_hocon34k_2024} as the Champion Challenger score \(S\) and shown in equation~\ref{eq:S}.

\begin{equation}\label{eq:MCC}
MCC =\frac{TP \cdot TN - FP \cdot FN}{\sqrt{P \cdot N \cdot (TP + FN)(TN + FP)}}
\end{equation}

\begin{equation}
\label{eq:M_norm}
M^{norm} = \frac{MCC + 1}{2}
\end{equation}

\begin{equation}
\label{eq:S}
S = \frac{M^{norm} + F_2}{2}
\end{equation}

\subsection{Experiments with GPT-4o}
\label{sec:GPT}
\begin{figure*}[!ht]
\begin{center}
\includegraphics[trim=0 80 0 70, clip, width=\textwidth]{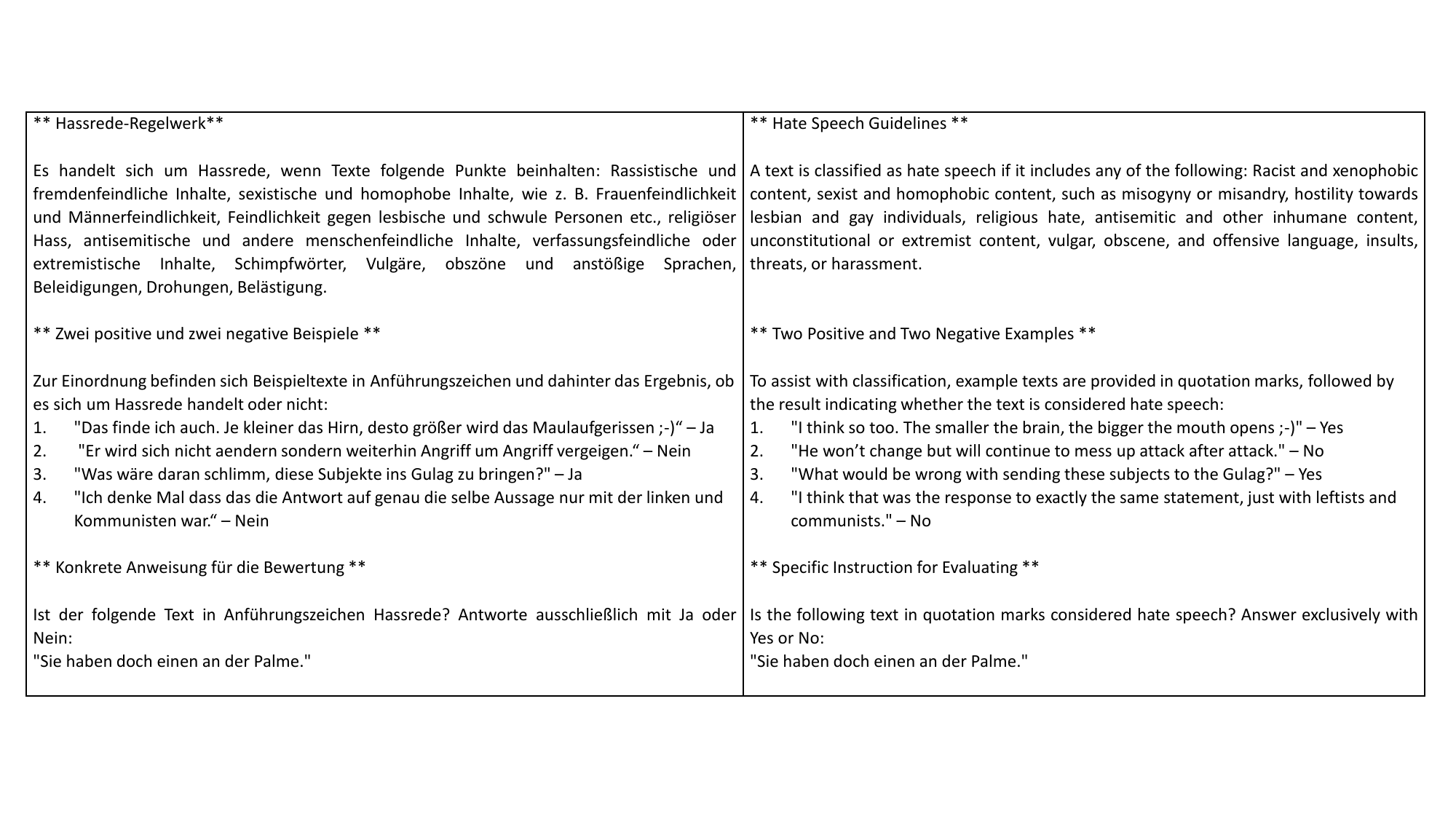} 
\caption{Prompt for GPT-4o Few-Shot Learning: Original Prompt (left) and Translation (right).}
\label{fig:prompt}
\end{center}
\end{figure*}

For the experiments with GPT-4, the GPT-4o model (exact designation: gpt-4o-2024-05-13) was used. 
In all GPT-4o experiments, the prompt first included the definition of hate speech as specified in Fig.~\ref{fig:prompt}. This definition was taken from \citet{keller_hocon34k_2024}.

Experiments are conducted with Zero-Shot, One-Shot, and Few-Shot learning. The prompt is structured into three parts. In the first part, the hate speech guidelines are explained. In the second part, for One-Shot learning, one labeled example text is provided, while for Few-Shot learning, four labeled example texts are given. The first Few-Shot learning example was also used for One-Shot learning. This part is omitted for Zero-Shot learning. The labels are either \textit{Yes} for hate speech or \textit{No} for no hate speech. The third part of the prompt is included in all experiments and contains the specific instruction. Only \textit{Yes} or \textit{No} are allowed as responses. Figure~\ref{fig:prompt} shows the prompt used for Few-Shot learning with four text examples, each labeled as \textit{Yes} or \textit{No}. The texts were taken from the HOCON34k training dataset (not the test dataset) and include original spelling errors.

API requests were made using the \texttt{openai.Completion.create} method from OpenAI's Python library \citep{openai_openai_2024}. The temperature parameter, ranging from 0.0 to 2.0, controls response creativity, with higher values producing more creative outputs. At temperature=0.0, responses are intended to be almost deterministic. For all experiments, the default value of 1.0 was used.
Preliminary tests showed no increase in determinism with a temperature of 0.0. The method output, either \textit{Yes} or \textit{No}, suggests that temperature has no impact on short responses.

\subsection{Experiments with Moderation API}
\label{sec:ModAPI}
The OpenAI Moderation API provides an endpoint to analyze input text to detect harmful content \citep{openai_openai_2024}. The underlying model is not publicly available and cannot be modified, retrained, or fine-tuned. In our experiments, we used the API's stable version from June 2024.
The \texttt{openai-Moderation.create} method allows client applications to check a text by passing it as an input parameter, along with a pre-generated API key. The service is paid, but offers free usage credits \citep{openai_openai_2024}. 

The API can process text up to 32,768 tokens and checks for harmful content across eleven categories: Sexual, Hate, Harassment, Self-Harm, Sexual/Minors, Hate/Threatening, Violence/Graphic, Self-Harm/Intent, Self-Harm/Instructions, Harassment/Threatening, and Violence \citep{openai_openai_2024}. This broad scope of content moderation aligns with, and in some cases exceeds, our own definition of hate speech, covering self-harm, instructions for self-harm, and graphic content involving death, violence, or injury. However, it does not explicitly cover unconstitutional or legal aspects, though these might fall under certain categories. Legal nuances specific to individual countries are not considered by the API.
The output of a \texttt{create()} call is a moderation object, with each category marked as \texttt{True} or \texttt{False} and a corresponding score. Additionally, an overall flag (\texttt{flagged}) is set to \texttt{True} if any category is marked as \texttt{True}. 
In our experiments, we focused solely on this overall flag and interpreted it as a binary classification for hate speech (\texttt{flagged == True} $\rightarrow 1$) or non-hate speech (\texttt{flagged == False} $\rightarrow 0$). More detailed classifications, such as sexism, were not further analyzed in our experiments.
The Moderation API does not use the \texttt{temperature} parameter for controlling randomness and creativity, and, according to OpenAI \citep{openai_openai_2024}, it provides nearly deterministic outputs. However, our experiments showed that the outputs were not always repeatable and exhibited slight variations.

\subsection{Experiments with Perspective API}
\label{sec:PersAPI}
\begin{table*}[htb]
\centering
\caption{Classification Results Across All Experiments and Comparison with Baseline}
\setlength{\tabcolsep}{4pt} 
\renewcommand{\arraystretch}{1.2} 
\small 
\begin{tabular}{|l|c|c|c|c|c|c|c|c|}
\hline
\textbf{Metric} & \multicolumn{3}{c|}{\textbf{GPT-4o}} & \multicolumn{3}{c|}{\textbf{Pers-API}} & \textbf{ModAPI} & \textbf{HOCON34k} \\
\cline{2-7}
       & \textbf{Zero-Shot} & \textbf{One-Shot} & \textbf{Few-Shot} & \(\boldsymbol{\tau = 0.38}\) & \(\boldsymbol{\tau = 0.5}\) & \(\boldsymbol{\tau = 0.8}\) & & \(\boldsymbol{\tau = 0.523}\) \\
\hline
Accuracy  & \underline{0.8049} & 0.7701 & 0.7988 & 0.7726 & \textbf{\underline{0.8097}} & 0.7965 & 0.6981 & 0.607 \\
Precision & \underline{0.5212} & 0.4660 & 0.5097 & 0.4516 & 0.5883 & \textbf{\underline{0.8571}}& 0.3766 & 0.327 \\
Recall    & 0.6849 & \underline{0.7700} & 0.6920 & \underline{0.4680} & 0.2766 & 0.0182 & 0.7031 & \textbf{0.851} \\
F1-score  & \textbf{\underline{0.5919}} & 0.5806 & 0.5870 & \underline{0.4597} & 0.3753 & 0.0357 & 0.4905 & 0.472 \\
F2-score  & 0.6444 &\textbf{\underline{0.6811}} & 0.6458 & \underline{0.4646} & 0.3091 & 0.0227 & 0.5992 & 0.644 \\
MCC       & \textbf{\underline{0.4743}} & 0.4612 & 0.4674 & \underline{0.3158} & 0.3066 & 0.1067 & 0.3326 & 0.320 \\
S         & 0.6908 &\textbf{\underline{0.7059}} & 0.6897 & \underline{0.5613} & 0.4812 & 0.2880 & 0.6327 & 0.652 \\
\hline
\end{tabular}
\vspace{0.4em}
\par \raggedright
Higher is better for all metrics. \textbf{Bold:} best overall value for each metric. \underline{Underlined:} best within the same model type.
\label{tab:results_1}
\end{table*}

The current generation of the Perspective API is implemented as a Charformer-based Transformer, referred to as Unified Toxic Content Classification (UTC) according to \citet{lees_new_2022}. 

A Charformer uses Gradient-Based Subword Tokenization (GBST), enabling the model to learn latent subwords from individual characters of a text, as explained by \citet{tay_charformer_2021}. The Perspective API model is pretrained on multilingual texts and comment sections \citep{lees_new_2022, tay_charformer_2021}. The pretrained model is not specifically focused on hate speech detection but rather on identifying toxic content.
Google Jigsaw defines toxicity as \textit{“[...] a rude, disrespectful, or unreasonable comment that is likely to make someone leave a discussion”} \citep{google_jigsaw_perspective_2024}.
Similar to the Moderation API, users cannot modify or fine-tune the underlying language model in the Perspective API.
Unlike the Moderation API, the Perspective API's response provides probability scores between 0 and 1 for the categories Toxicity, Severe Toxicity, Identity Attack, Insult, Profanity, and Threat. Currently, no assessment for sexist texts is provided \citep{perspective_perspective_2024}, but except for this, the definition is similar to the one used in \citet{keller_hocon34k_2024}. Legal or unconstitutional violations are not considered, as these are country-specific issues, and legal violations such as incitement to hatred in Germany are also not accounted for. Legal intricacies of individual countries are not part of the Perspective API’s evaluation.

In our experiments, we mapped the output scores to binary values, requiring the selection of a threshold. For research purposes, a threshold of 0.7 to 0.9 is recommended for the Perspective API, with adjustments based on the specific application \citep{perspective_perspective_2024}. A separate threshold can be set for each category. In our experiments, we used thresholds \( \tau \in \{0.38, 0.5, 0.8\} \) across all categories. If the score for any category exceeds the threshold, the text is classified as hate speech (1); otherwise, it is classified as non-hate speech (0).


\pagebreak
We conducted three experiments with the Perspective API using the specified thresholds, incorporating waiting times between requests to avoid exceeding the quota.
For Python requests, we used the Google API Client Library (\texttt{google-api-python-client}). A request is made using the \texttt{execute} method, with input parameters including the text to be evaluated, the categories to be checked, and the target language. The response is returned as a JSON object, containing a score for each category. The highest score from the response is used to determine the result, which is then mapped to \mbox{hate speech (1)} or non-hate \mbox{speech (0)} based on the selected threshold.

\section{Results and Limitations}
\label{sec:results}

\subsection{Results}

The classification results of the experiments with GPT-4o and the HOCON34k rule set show no significant differences between Zero-Shot, One-Shot, and Few-Shot learning. The approaches differ only slightly. Zero-Shot learning, without any example texts, achieved the best results in terms of Accuracy, Precision, F1-score, and MCC. One-Shot learning yielded the highest scores for Recall, F2-score, and Champion-Challenger score (\(S\)). A comparison with the baseline from the BERT-based HOCON34k model \citep{keller_hocon34k_2024} shows that GPT-4o in One-Shot learning achieved an improved \(S\) score, approximately 5 percentage points higher than the baseline (\(S\)=0.7059).
Contrary to our expectations, Few-Shot learning (with four example texts) performed slightly worse. 

Table~\ref{tab:results_1} presents the results of the experiments, where the average of three runs was calculated for each metric. For the Perspective API, a threshold of 0.38 was the most effective for detecting hate speech. A threshold of 0.5 produced better results for Accuracy, while a threshold of 0.8 yielded the best Precision. A comparison of Perspective and Moderation API shows that the Moderation API achieved superior overall performance on our test dataset. The high Accuracy and Precision scores of the Perspective API are due to mostly negative predictions combined with the dataset's imbalance. With \( \tau = 0.8 \), only 6 out of 1,592 samples were classified as hate speech, resulting in a high number of 323 False Negatives (see Tab.~\ref{tab:cm_1}).
For our application scenario, we primarily compare quality using the Champion-Challenger score (\(S\)). For the baseline model, the threshold of 0.523 was optimized for \(S\), effectively balancing the F2-score, MCC, and indirectly, Recall. This optimization is reflected in the Recall value of the HOCON34k baseline, which achieved the highest score of 0.851.
The optimization for \(S\) resulted in more True Positives being detected, but also significantly more False Positives (see Tab.~\ref{tab:cm_1}). 

However, there were considerably fewer False Negatives. The baseline classification aimed to detect as much hate speech as possible, even at the cost of a higher number of False Positives, ensuring that no hate speech remains undetected.
It should also be noted that the Perspective API, without language selection, had greater difficulty with German dialects, HTTP links, and uninterpretable character strings. For example, texts like “Des häd a Depp a gsogt!” (English translation: “That's what a fool said!”) or “mach mal den rand zu, du kleine braune drecksau!!!” (English translation: “close the rim, you little brown bastard!!!”) were classified as non-hate speech due to error messages from the Perspective API. However, when the language was explicitly set to German, no error messages were returned.
Table~\ref{tab:cm_1} also shows that deterministic output is not guaranteed for identical queries with GPT-4o and the Moderation API. There were slight variations between individual test runs, which was observed for both One-Shot and Few-Shot Learning with GPT-4o. In contrast, the output from the Perspective API appears to be repeatable, as demonstrated in Tab.~\ref{tab:cm_1} for the threshold of 0.8. This consistency was also observed in test runs with the other two thresholds.
\begin{table}[ht]
\centering
\caption{Confusion Matrices for 3 Repeated Runs}
\setlength{\tabcolsep}{4pt} 
\renewcommand{\arraystretch}{1.2} 
\small 
\begin{tabular}{|l|r|r|r|r|}
\hline
\textbf{Experiment} & \textbf{TP} & \textbf{FN} & \textbf{FP} & \textbf{TN} \\
\hline
GPT-4o, Few-Shot (1)   & 229 & 100 & 221 & 1042 \\
GPT-4o, Few-Shot (2)   & 227 & 102 & 218 & 1045 \\
GPT-4o, Few-Shot (3)   & 227 & 102 & 218 & 1045 \\
Moderation API (1)              & 231 & 98  & 384 & 879  \\
Moderation API (2)              & 231 & 98  & 382 & 881  \\
Moderation API (3)              & 232 & 97  & 383 & 880  \\
Pers-API, \( \tau = 0.8 \) (1) & 6   & 323 & 1   & 1262 \\
Pers-API, \( \tau = 0.8 \) (2) & 6   & 323 & 1   & 1262 \\
Pers-API, \( \tau = 0.8 \) (3) & 6   & 323 & 1   & 1262 \\
HOCON34k \( \tau = 0.523 \) & 280 & 49  & 577 & 686  \\
\hline
\end{tabular}
\label{tab:cm_1}
\end{table}

\subsection{Limitations}

The HOCON34k test dataset from \citet{keller_hocon34k_2024}, containing 1,592 text examples, has an uneven distribution with a significantly higher proportion of non-hate speech. The same applies to the HOCON34k training dataset, which was used to train the HOCON34k baseline. The underlying models and training datasets for the other solutions analyzed are not disclosed.
During the analysis of individual predictions, weaknesses in the test dataset were identified. Upon closer inspection, annotations were found that, in our opinion, do not comply with the given HOCON34k guidelines.  For example, the text  “stellt das schwein und macht es wie die leute in bolivien !” (English
translation: “put the pig and make it like the people in bolivia !”) was labeled as non hate speech, although it is hate speech according to the HOCON34 rules.
Eliminating potentially erroneous data might lead to different results.

Most of the models analyzed are non-deterministic and can produce different predictions for repeated runs of the classification. For the Perspective API, determinism is assumed. However, this assumption is based solely on the consistent classifications observed in the three repetitions conducted in our experiments. The assumption was not definitively proven in this study. In contrast, the BERT-based HOCON34k classifier produces consistent outputs for identical input texts on the same hardware, exhibiting deterministic behavior.

The specific definition of hate speech in HOCON34k does not align with the definitions used in the underlying models for the Perspective and Moderation APIs. Different definitions of hate speech or harmful speech for the various solutions limit their comparability. The models of the APIs analyzed were trained on harmful content using different definitions. However, aside from legal violations or unconstitutional statements, a high degree of overlap between the models is assumed. 

\label{sec:reannotation}
\begin{figure}[b]
\begin{center}
\includegraphics[trim=60 85 75 100, clip, width=\columnwidth]{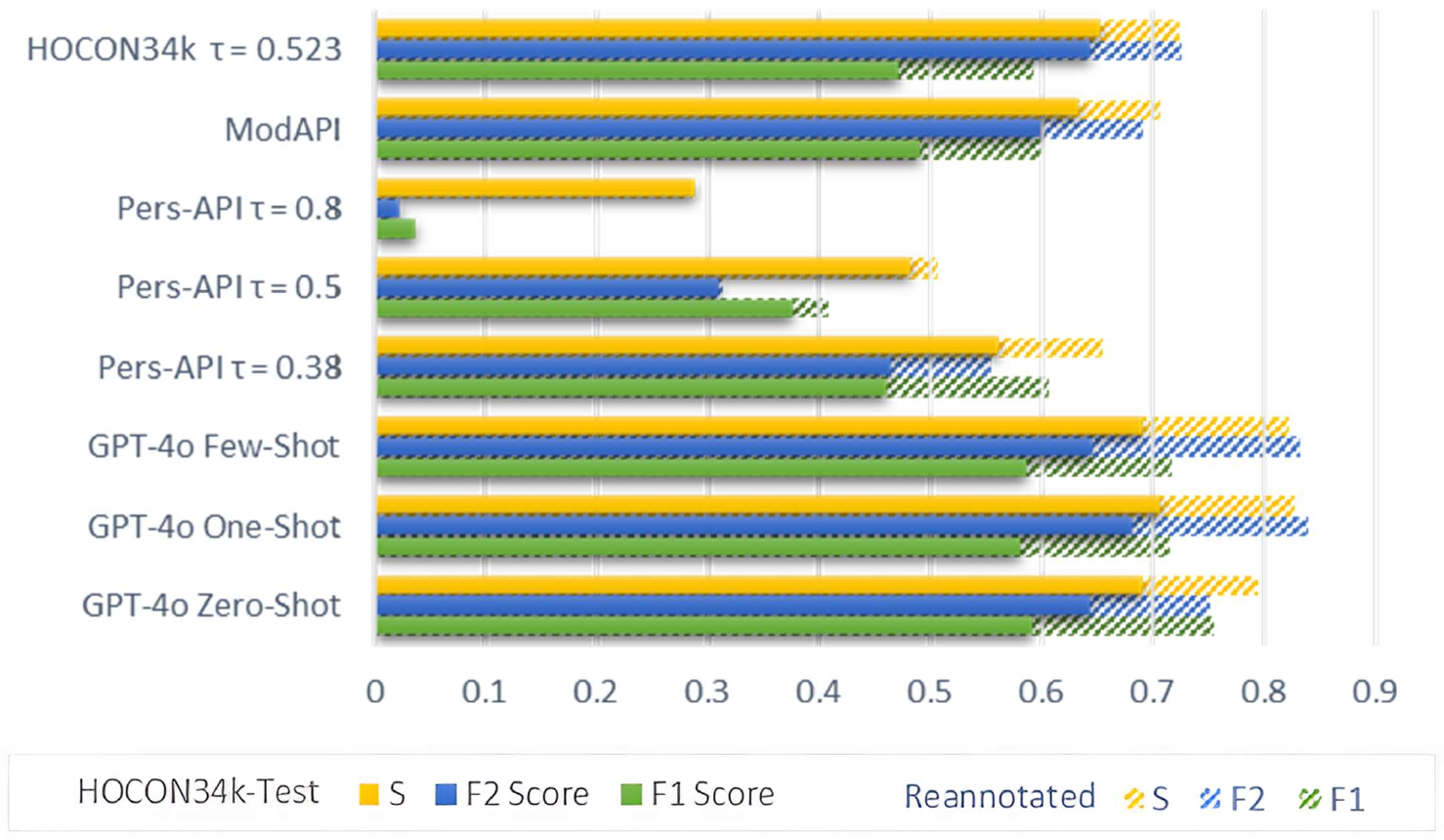}
\caption{S-, F1-, and F2-scores of all models on HOCON34k test compared to the reannotated data.}
\label{fig:results_both}
\end{center}
\end{figure}

\begin{table}[H]
\centering
\caption{Overview of Adjustments to the Test Dataset through Reannotation (HS = Hate Speech).}
\begin{tabular}{|l|r|}
\hline
Reviewed labels & 313 \\
\hline
{Unanimous decisions by annotators} & 155 \\
\hspace{5mm}\textit{in accordance with GPT-4o:} & \textit{107} \\
{Decisions without unanimity} & 158 \\
\hspace{5mm}\textit{in accordance with GPT-4o:} & \textit{94} \\
\hline
{Samples where the label was changed} & 201 \\
 \hspace{5mm}From HS to Not HS: & 64 \\
 \hspace{5mm}From Not HS to HS: & 137 \\
\hline
Samples without change & 112 \\
\hline
\end{tabular}
\label{tab:changes}
\end{table}

\begin{table*}[ht]
\centering
\caption{Classification Results Across All Experiments after Reannotation}
\setlength{\tabcolsep}{4pt} 
\renewcommand{\arraystretch}{1.2} 
\small 
\begin{tabular}{|l|c|c|c|c|c|c|c|c|}
\hline
\textbf{Metric} & \multicolumn{3}{c|}{\textbf{GPT-4o}} & \multicolumn{3}{c|}{\textbf{Pers-API}} & \textbf{ModAPI} & \textbf{HOCON34k} \\
\cline{2-7}
       & \textbf{Zero-Shot} & \textbf{One-Shot} & \textbf{Few-Shot} & \(\boldsymbol{\tau=0.38}\) & \(\boldsymbol{\tau=0.5}\) & \(\boldsymbol{\tau=0.8}\) & & \(\boldsymbol{\tau=0.523}\) \\
\hline
Accuracy  & \textbf{\underline{0.8770}} & 0.8102 & 0.8139 & \underline{0.8290} & 0.8020 & 0.7517 & 0.7427 & 0.703 \\
Precision & \underline{0.7592} & 0.5748 & 0.5817 & 0.7216 & 0.8258 & \textbf{\underline{1.0000}} & 0.4932 & 0.453 \\
Recall    & 0.7498 & \textbf{\underline{0.9493}} & 0.9327 & \underline{0.5237} & 0.2718 & 0.0150 & 0.7672 & 0.855 \\
F1-score & \textbf{\underline{0.7545}} & 0.7160 & 0.7165 & \underline{0.6069} & 0.4090 & 0.0295 & 0.6005 & 0.592 \\
F2-score & 0.7516 & \textbf{\underline{0.8398}} & 0.8322 & \underline{0.5541} & 0.3139 & 0.0186 & 0.6906 & 0.726 \\
MCC       & \textbf{\underline{0.6725}} & 0.6277 & 0.6252 & \underline{0.5117} & 0.3974 & 0.1060 & 0.4461 & 0.440 \\
S         & 0.7939 & \textbf{\underline{0.8268}} & 0.8224 & \underline{0.6550} & 0.5063 & 0.2858 & 0.7068 & 0.723 \\
\hline
\end{tabular}
\vspace{0.4em}
\par \raggedright
Higher is better for all metrics. \textbf{Bold:} best overall value for each metric. \underline{Underlined:} best within the same model type.
\label{tab:results_2}
\end{table*}

\begin{table}[ht]
\centering
\caption{Confusion Matrices after Reannotation}
\setlength{\tabcolsep}{4pt} 
\renewcommand{\arraystretch}{1.2} 
\small 
\begin{tabular}{|l|r|r|r|r|}
\hline
\textbf{Experiment} & \textbf{TP} & \textbf{FN} & \textbf{FP} & \textbf{TN} \\
\hline
GPT-4o, Few-Shot (1)   & 376 & 25  & 279 & 911 \\
GPT-4o, Few-Shot (2)   & 375 & 26  & 262 & 928 \\
GPT-4o, Few-Shot (3)   & 371 & 30  & 266 & 924 \\
Moderation API (1)              & 308 & 93  & 308 & 872 \\
Moderation API (2)              & 308 & 93  & 315 & 875 \\
Moderation API (3)              & 307 & 94  & 315 & 875 \\
Pers-API, \( \tau = 0.8 \) (1) & 6   & 395 & 0   & 1190 \\
Pers-API, \( \tau = 0.8 \) (2) & 6   & 395 & 0   & 1190 \\
Pers-API, \( \tau = 0.8 \) (3) & 6   & 395 & 0   & 1190 \\
HOCON34k \( \tau = 0.557 \) & 343 & 58  & 415 & 775  \\
\hline
\end{tabular}
\label{tab:cm_2}
\end{table}

\section{Dataset Analysis and Improvement}
\subsection{Analysis of Testdata}

Due to the assumption of optimization potential in the dataset, a reannotation of a sample of 91 text examples from the test dataset was conducted by four additional individuals who had previously worked on the HOCON34k dataset and were therefore well-versed in the guidelines. In the reannotation process, between 16 and 26 text examples were annotated differently compared to the test dataset. In some cases, it was not possible to make an exact assessment based solely on the text. The four annotators were thus asked to indicate whether the context was sufficient for an accurate assessment. Among the 91 text examples, 18, 21, 42, and 45 texts were marked as "Insufficient Context" by the four annotators. The annotators agreed on the annotation of 51 texts. In 5 out of 91 text examples, all annotators agreed that the annotation in the dataset was incorrect.
A projection for the entire test dataset of 1,592 samples estimated a range of 78 to 96 samples with potentially incorrect annotations (error range 10\,\%, confidence interval 90\,\%). The experiments with GPT-4o and Few-Shot Learning yielded the same results as the annotators in cases of unanimous agreement. For the 51 unanimously confirmed texts, GPT-4o achieved an accuracy of 100\,\%. Based on these results, it is assumed that the imperfect classification is not solely due to model limitations but also to inconsistent or erroneous annotations.

\subsection{Reannotation of Testdata}

A reannotation of the HOCON34k test dataset was conducted according to the following procedure. To keep the manual annotation effort within a moderate scope, automated classifications based on GPT-4o were used to identify potential misclassifications, which were then manually reviewed only when necessary. For this, three classification runs were performed for all test dataset samples using the Zero-Shot version of GPT-4o, and a majority decision was formed based on the mode value.

The majority decisions generated by GPT-4o were then compared with the labels present in the test dataset. If there was a discrepancy between the comparison values, the corresponding samples were manually rechecked according to the HOCON34k guidelines. Following this process, 314 labels requiring review were identified. 
These were subsequently reannotated by three annotators independently in a blind annotation process. The labels in the test dataset were then adjusted, with the majority decision of the three annotators being adopted as the new label. In about 64\% of the cases, the label suggested by GPT was confirmed by the annotators and incorporated into the dataset as the new label. In the remaining cases, the majority of annotators disagreed with the GPT-generated label, and the original label was retained. Due to a missing annotation for one of the samples under review, it was removed from the dataset. 

This results in a new dataset size of 1,591 samples. Among these, 401 are labeled as hate speech (previously 329), and 1,190 are labeled as non-hate speech (previously 1,263). Compared to the original dataset, the proportion of hate speech has increased significantly from 20.7\,\% to 25.2\,\%.
Table~\ref{tab:changes} summarizes the reannotation process and adjustments made to the test dataset.

\subsection{Improved Results}

Following the reannotation described earlier, a replication of all previously conducted experiments was performed using the resulting optimized test dataset. Additionally, the test of the HOCON34k baseline classifier was repeated on the new test data.
The results of the repeated experiments are presented in Tab.~\ref{tab:results_2} and Fig.~\ref{fig:results_both}. Tab.~\ref{tab:cm_2} shows the confusion matrices for the repeated experiments. The test results improved for most experiments after reannotation, often showing a significant increase in performance.

Overall, the GPT-4o variants demonstrated the best results on both the original and reannotated test datasets. In contrast to the previous results, GPT-4o also outperformed the other models in terms of accuracy and recall. Based on the reannotated test dataset, with the exception of Zero-Shot learning the GPT-based classification variants achieved F2-scores and \(S\)-scores above 0.8, outperforming the other models in these and most other metrics. The best result was achieved by the One-Shot variant with \(S\) = 0.8268. This variant already performed best in the original experiments and experienced an increase in the \(S\)-score by 17.13\,\% due to the reannotation.
The reannotation also improved the test results for the HOCON34k baseline classifier \citep{keller_hocon34k_2024}. The \(S\)-score increased from 0.652 to 0.723, a 10.89\,\% improvement. The GPT-based variants experienced a stronger increase in the \(S\)-score. The One-Shot variant nearly doubled its advantage over the baseline, with performance increasing from 0.054 to 0.104. The improvement in recall was even more significant. While the baseline classifier initially demonstrated the highest recall, the One-Shot variant achieved a substantial improvement of 23.29\,\%, increasing from 0.77 to 0.9493, thereby surpassing the baseline also in recall performance. The recall for the baseline classifier, on the other hand, remained mostly unchanged.
The Perspective API, measured by \(S\)-score and F2-score, again produced the weakest results, in some cases even showing a decline compared to the previous evaluation with the original dataset. The Moderation API achieved better results than the Perspective API but remained significantly behind GPT-4o and slightly behind the HOCON34k baseline. Figure~\ref{fig:results_both} provides a comparison of the test results before and after the reannotation of the test data. The graph illustrates the significant performance improvements in nearly all experiments. The largest performance gain was observed with GPT-4o. However, this should be viewed critically, as the reannotation process only considered samples that GPT identified as misclassified. Given the near-deterministic classification, each adjustment results in a predictable improvement in evaluation outcomes for GPT. Although the final decision to adjust labels was made by annotators, independent of GPT's assessments, the selective review may introduce a bias in favor of GPT.

\balance

\section{Conclusions}
\label{sec:conclusions}
In this study, we compared GPT-4o, OpenAI's Moderation API, and the Perspective API in detecting hate speech in German online comments, using the HOCON34k dataset and its baseline classifier as a reference. GPT-4o, using various prompting strategies, outperformed other models, with One-Shot Learning yielding better results than Few-Shot Learning. The Moderation API performed well, while the Perspective API struggled, showing high precision but missing most hate comments due to numerous false negatives.

Our findings highlight the critical role of high-quality datasets in improving classification performance. Correcting annotation errors resulted in over 10\% improvement across most models, including the HOCON34k classifier. 


Future research should focus on incorporating contextual information into hate speech detection models, as expressions often depend on previous comments or the articles \citep{madhu_detecting_2023}. Expanding datasets through data augmentation \citep{jahan_comprehensive_2024} and developing systems capable of continuous learning, which adapt to evolving language using moderator feedback, are also essential. Specialized models for different sections of online newspapers, like sports or politics, could further improve detection accuracy. These advancements are a key to enhancing real-world hate speech detection systems.

\nocite{*}
\section*{References}\label{sec:reference}

\bibliographystyle{lrec-coling2024-natbib}
\bibliography{Paper-GPT}

\begin{thebibliography}{34}
\expandafter\ifx\csname natexlab\endcsname\relax\def\natexlab#1{#1}\fi

\bibitem[{Alkomah and Ma(2022)}]{alkomah_literature_2022}
Fatimah Alkomah and Xiaogang Ma. 2022.
\newblock \href {https://doi.org/10.3390/info13060273} {A {Literature} {Review} of {Textual} {Hate} {Speech} {Detection} {Methods} and {Datasets}}.
\newblock \emph{Information}, 13(6):273.

\bibitem[{Chan et~al.(2020)Chan, Schweter, and Möller}]{chan_germans_2020}
Branden Chan, Stefan Schweter, and Timo Möller. 2020.
\newblock \href {http://arxiv.org/abs/2010.10906} {German's {Next} {Language} {Model}}.

\bibitem[{Chicco and Jurman(2020)}]{chicco_advantages_2020}
Davide Chicco and Giuseppe Jurman. 2020.
\newblock \href {https://doi.org/10.1186/s12864-019-6413-7} {The advantages of the {Matthews} correlation coefficient ({MCC}) over {F1} score and accuracy in binary classification evaluation}.
\newblock \emph{BMC Genomics}, 21(1):6.

\bibitem[{Chiu et~al.(2021)Chiu, Collins, and Alexander}]{chiu_detecting_2021}
Ke-Li Chiu, Annie Collins, and Rohan Alexander. 2021.
\newblock \href {https://doi.org/10.48550/ARXIV.2103.12407} {Detecting {Hate} {Speech} with {GPT}-3}.

\bibitem[{Devlin et~al.(2019)Devlin, Chang, Lee, and Toutanova}]{devlin_bert_2019}
Jacob Devlin, Ming-Wei Chang, Kenton Lee, and Kristina Toutanova. 2019.
\newblock \href {http://arxiv.org/abs/1810.04805} {{BERT}: {Pre}-training of {Deep} {Bidirectional} {Transformers} for {Language} {Understanding}}.

\bibitem[{{ERIC}(2016)}]{eric_europaische_2016}
{ERIC}. 2016.
\newblock \href {https://rm.coe.int/ecri-general-policy/recommendation-no-15-on-combating-hatespeech-germ/16808b5b00} {Europäische {Kommission} gegen {Rassismus} und {Intoleranz} {Allgemeine} {Politik}-{Empfehlung} {Nr}. 15 der {ECRI} über die {Bekämpfung} von {Hassrede}}.

\bibitem[{{European Parliament}(2022)}]{european_parliament_verordnung_2022}
{European Parliament}. 2022.
\newblock \href {https://eur-lex.europa.eu/legal-content/DE/TXT/?uri=CELEX%3A32022R2065} {Verordnung - 2022/2065 - {EN} - {EUR}-{Lex}}.

\bibitem[{Glasebach et~al.(2024)Glasebach, Keller, Döschl, and Mandl}]{glasebach_gmhp7k_2024}
Jonas Glasebach, Max-Emanuel Keller, Alexander Döschl, and Peter Mandl. 2024.
\newblock \href {https://doi.org/10.1609/icwsm.v18i1.31438} {{GMHP7k}: {A} {Corpus} of {German} {Misogynistic} {Hatespeech} {Posts}}.
\newblock \emph{Proceedings of the International AAAI Conference on Web and Social Media}, 18:1946--1957.

\bibitem[{{Google Jigsaw}(2024)}]{google_jigsaw_perspective_2024}
{Google Jigsaw}. 2024.
\newblock \href {https://www.perspectiveapi.com/} {Perspective {API}}.
\newblock [Accessed: 2024-09-21].

\bibitem[{Guo et~al.(2024)Guo, Hu, Mu, Shi, Zhao, Vishwamitra, and Hu}]{guo_investigation_2024}
Keyan Guo, Alexander Hu, Jaden Mu, Ziheng Shi, Ziming Zhao, Nishant Vishwamitra, and Hongxin Hu. 2024.
\newblock \href {https://doi.org/10.48550/ARXIV.2401.03346} {An {Investigation} of {Large} {Language} {Models} for {Real}-{World} {Hate} {Speech} {Detection}}.

\bibitem[{Hosseini et~al.(2017)Hosseini, Kannan, Zhang, and Poovendran}]{hosseini_deceiving_2017}
Hossein Hosseini, Sreeram Kannan, Baosen Zhang, and Radha Poovendran. 2017.
\newblock \href {https://doi.org/10.48550/ARXIV.1702.08138} {Deceiving {Google}'s {Perspective} {API} {Built} for {Detecting} {Toxic} {Comments}}.

\bibitem[{Istaiteh et~al.(2020)Istaiteh, Al-Omoush, and Tedmori}]{istaiteh_racist_2020}
Othman Istaiteh, Razan Al-Omoush, and Sara Tedmori. 2020.
\newblock \href {https://doi.org/10.1109/IDSTA50958.2020.9264052} {Racist and {Sexist} {Hate} {Speech} {Detection}: {Literature} {Review}}.
\newblock In \emph{2020 {International} {Conference} on {Intelligent} {Data} {Science} {Technologies} and {Applications} ({IDSTA})}, pages 95--99.

\bibitem[{Jahan and Oussalah(2023)}]{jahan_systematic_2023}
Md~Saroar Jahan and Mourad Oussalah. 2023.
\newblock \href {https://doi.org/10.1016/j.neucom.2023.126232} {A systematic review of hate speech automatic detection using natural language processing}.
\newblock \emph{Neurocomputing}, 546:126232.

\bibitem[{Jahan et~al.(2024)Jahan, Oussalah, Beddia, Mim, and Arhab}]{jahan_comprehensive_2024}
Md~Saroar Jahan, Mourad Oussalah, Djamila~Romaissa Beddia, Jhuma~kabir Mim, and Nabil Arhab. 2024.
\newblock \href {http://arxiv.org/abs/2404.00303} {A {Comprehensive} {Study} on {NLP} {Data} {Augmentation} for {Hate} {Speech} {Detection}: {Legacy} {Methods}, {BERT}, and {LLMs}}.

\bibitem[{Jaki and Steiger(2023)}]{jaki_hate_2023}
Sylvia Jaki and Stefan Steiger. 2023.
\newblock \href {https://doi.org/10.1007/978-3-662-65964-9_1} {Hate {Speech} online: {Hartknäckiges} {Phänomen} und interdisziplinärer {Forschungsgegenstand}}.
\newblock In Sylvia Jaki and Stefan Steiger, editors, \emph{Digitale {Hate} {Speech}}, pages 1--14. Springer Berlin Heidelberg.

\bibitem[{Johnson and Khoshgoftaar(2019)}]{johnson_survey_2019}
Justin~M. Johnson and Taghi~M. Khoshgoftaar. 2019.
\newblock \href {https://doi.org/10.1186/s40537-019-0192-5} {Survey on deep learning with class imbalance}.
\newblock \emph{Journal of Big Data}, 6(1):27.

\bibitem[{Keller et~al.(2024)Keller, Auch, Döschl, Vlk, Quernheim, Hartmann, Mandl, Kaul, and Franz}]{keller_hocon34k_2024}
Max-Emanuel Keller, Maximilian Auch, Alexander Döschl, Fabian Vlk, Julian Quernheim, Mike Hartmann, Peter Mandl, Alexander Kaul, and Markus Franz. 2024.
\newblock \href {https://doi.org/10.1007/978-3-031-78090-5_18} {{HOCON34k}: {A} {Corpus} of {Hate} speech in {Online} {Comments} from {German} {Newspapers}}.
\newblock In \emph{Information {Integration} and {Web} {Intelligence}}, Bratislava, Slovakia.

\bibitem[{Lees et~al.(2022)Lees, Tran, Tay, Sorensen, Gupta, Metzler, and Vasserman}]{lees_new_2022}
Alyssa Lees, Vinh~Q. Tran, Yi~Tay, Jeffrey Sorensen, Jai Gupta, Donald Metzler, and Lucy Vasserman. 2022.
\newblock \href {https://doi.org/10.48550/ARXIV.2202.11176} {A {New} {Generation} of {Perspective} {API}: {Efficient} {Multilingual} {Character}-level {Transformers}}.

\bibitem[{Li et~al.(2023)Li, Fan, Atreja, and Hemphill}]{li_hot_2023}
Lingyao Li, Lizhou Fan, Shubham Atreja, and Libby Hemphill. 2023.
\newblock \href {https://doi.org/10.48550/ARXIV.2304.10619} {"{HOT}" {ChatGPT}: {The} promise of {ChatGPT} in detecting and discriminating hateful, offensive, and toxic comments on social media}.

\bibitem[{Madhu et~al.(2023)Madhu, Satapara, Modha, Mandl, and Majumder}]{madhu_detecting_2023}
Hiren Madhu, Shrey Satapara, Sandip Modha, Thomas Mandl, and Prasenjit Majumder. 2023.
\newblock \href {https://doi.org/10.1016/j.eswa.2022.119342} {Detecting offensive speech in conversational code-mixed dialogue on social media: {A} contextual dataset and benchmark experiments}.
\newblock \emph{Expert Systems with Applications}, 215:119342.

\bibitem[{Mandl(2023)}]{jaki_ki-verfahren_2023}
Thomas Mandl. 2023.
\newblock \href {https://doi.org/10.1007/978-3-662-65964-9_6} {{KI}-{Verfahren} für die {Hate} {Speech} {Erkennung}: {Die} {Gestaltung} von {Ressourcen} für das maschinelle {Lernen} und ihre {Zuverlässigkeit}}.
\newblock In Sylvia Jaki and Stefan Steiger, editors, \emph{Digitale {Hate} {Speech}}, pages 111--130. Springer Berlin Heidelberg.

\bibitem[{Markov et~al.(2022)Markov, Zhang, Agarwal, Eloundou, Lee, Adler, Jiang, and Weng}]{markov_holistic_2022}
Todor Markov, Chong Zhang, Sandhini Agarwal, Tyna Eloundou, Teddy Lee, Steven Adler, Angela Jiang, and Lilian Weng. 2022.
\newblock \href {https://doi.org/10.48550/ARXIV.2208.03274} {A {Holistic} {Approach} to {Undesired} {Content} {Detection} in the {Real} {World}}.

\bibitem[{Matter et~al.(2024)Matter, Schirmer, Grinberg, and Pfeffer}]{matter_close_2024}
Daniel Matter, Miriam Schirmer, Nir Grinberg, and Jürgen Pfeffer. 2024.
\newblock \href {https://doi.org/10.48550/ARXIV.2401.02001} {Close to {Human}-{Level} {Agreement}: {Tracing} {Journeys} of {Violent} {Speech} in {Incel} {Posts} with {GPT}-4-{Enhanced} {Annotations}}.

\bibitem[{{Meta}(2024)}]{meta_hassrede_2024}
{Meta}. 2024.
\newblock \href {https://transparency.meta.com/de-de/policies/community-standards/hate-speech/} {Hassrede {\textbar} {Transparency} {Center}}.
\newblock [Accessed: 2024-09-21].

\bibitem[{Nogara et~al.(2023)Nogara, Pierri, Cresci, Luceri, Törnberg, and Giordano}]{nogara_toxic_2023}
Gianluca Nogara, Francesco Pierri, Stefano Cresci, Luca Luceri, Petter Törnberg, and Silvia Giordano. 2023.
\newblock \href {https://doi.org/10.48550/ARXIV.2312.12651} {Toxic {Bias}: {Perspective} {API} {Misreads} {German} as {More} {Toxic}}.

\bibitem[{{OpenAI}(2023)}]{openai_gpt-4_2023}
{OpenAI}. 2023.
\newblock \href {https://doi.org/10.48550/ARXIV.2303.08774} {{GPT}-4 {Technical} {Report}}.
\newblock [Accessed: 2024-09-21].

\bibitem[{{OpenAI}(2024)}]{openai_openai_2024}
{OpenAI}. 2024.
\newblock \href {https://platform.openai.com} {{OpenAI} {Platform}}.
\newblock [Accessed: 2024-09-21].

\bibitem[{Pan et~al.(2024)Pan, García-Díaz, and Valencia-García}]{pan_comparing_2024}
Ronghao Pan, José~Antonio García-Díaz, and Rafael Valencia-García. 2024.
\newblock \href {https://doi.org/10.32604/cmes.2024.049631} {Comparing {Fine}-{Tuning}, {Zero} and {Few}-{Shot} {Strategies} with {Large} {Language} {Models} in {Hate} {Speech} {Detection} in {English}}.
\newblock \emph{Computer Modeling in Engineering \& Sciences}, 140(3):2849--2868.

\bibitem[{{Perspective}(2024)}]{perspective_perspective_2024}
{Perspective}. 2024.
\newblock \href {https://developers.perspectiveapi.com/s/?language=en_US} {Perspective {\textbar} {Developers}}.
\newblock [Accessed: 2024-09-21].

\bibitem[{Powers(2020)}]{powers_evaluation_2020}
David~MW Powers. 2020.
\newblock \href {https://arxiv.org/abs/2010.16061} {Evaluation: from precision, recall and {F}-measure to {ROC}, informedness, markedness and correlation}.
\newblock \emph{arXiv:2010.16061}.

\bibitem[{Rawat et~al.(2024)Rawat, Kumar, and Samant}]{rawat_hate_2024}
Anchal Rawat, Santosh Kumar, and Surender~Singh Samant. 2024.
\newblock \href {https://doi.org/10.1002/wics.1648} {Hate speech detection in social media: {Techniques}, recent trends, and future challenges}.
\newblock \emph{WIREs Computational Statistics}, 16(2):e1648.

\bibitem[{Tay et~al.(2021)Tay, Tran, Ruder, Gupta, Chung, Bahri, Qin, Baumgartner, Yu, and Metzler}]{tay_charformer_2021}
Yi~Tay, Vinh~Q. Tran, Sebastian Ruder, Jai Gupta, Hyung~Won Chung, Dara Bahri, Zhen Qin, Simon Baumgartner, Cong Yu, and Donald Metzler. 2021.
\newblock \href {https://doi.org/10.48550/ARXIV.2106.12672} {Charformer: {Fast} {Character} {Transformers} via {Gradient}-based {Subword} {Tokenization}}.

\bibitem[{Udanor and Anyanwu(2019)}]{udanor_combating_2019}
Collins Udanor and Chinatu~C. Anyanwu. 2019.
\newblock \href {https://doi.org/10.1108/DTA-01-2019-0007} {Combating the challenges of social media hate speech in a polarized society: {A} {Twitter} ego lexalytics approach}.
\newblock \emph{Data Technologies and Applications}, 53(4):501--527.

\bibitem[{Vaswani et~al.(2017)Vaswani, Shazeer, Parmar, Uszkoreit, Jones, Gomez, Kaiser, and Polosukhin}]{vaswani_attention_2017}
Ashish Vaswani, Noam Shazeer, Niki Parmar, Jakob Uszkoreit, Llion Jones, Aidan~N. Gomez, Lukasz Kaiser, and Illia Polosukhin. 2017.
\newblock \href {http://arxiv.org/abs/1706.03762} {Attention {Is} {All} {You} {Need}}.

\end{thebibliography}


\end{document}